\documentclass[letterpaper, 10 pt, conference]{ieeeconf}  

\IEEEoverridecommandlockouts                             
\usepackage{amsmath,amssymb,amsfonts}
\usepackage{algorithmic}
\usepackage{graphicx}
\usepackage{textcomp}
\usepackage{xcolor}
\usepackage{algorithm}
\usepackage{array}
\usepackage[caption=false,font=normalsize,labelfont=sf,textfont=sf]{subfig}
\usepackage{stfloats}
\usepackage{url}
\usepackage{verbatim}
\usepackage{times}
\usepackage{epsfig}
\usepackage{float}
\usepackage{wrapfig}
\usepackage{bm,xspace}
\usepackage{comment}
\usepackage{multirow}
\usepackage{balance}
\usepackage{booktabs}
\usepackage{etoolbox,siunitx}
\usepackage{calc}
\usepackage{pifont,hologo}
\usepackage{nicefrac}
\usepackage{makecell}
\usepackage{svg}
\usepackage{enumerate}
\usepackage{adjustbox}
\usepackage[colorlinks=black,citecolor=green]{hyperref}

\usepackage{amsmath,amsfonts,bm}

\def\Figref#1{Fig.~\ref{#1}}

\def\eqref#1{equation~\ref{#1}}

\def\1{\bm{1}}

\DeclareMathAlphabet{\mathsfit}{\encodingdefault}{\sfdefault}{m}{sl}
\SetMathAlphabet{\mathsfit}{bold}{\encodingdefault}{\sfdefault}{bx}{n}

\title{\LARGE \bf
Camera Relocalization in Shadow-free Neural Radiance Fields
}

\author{Shiyao Xu$^{1,*}$, Caiyun Liu$^{1,*}$, Yuantao Chen$^{1,2}$, Zhenxin Zhu$^{3}$, Zike Yan$^{1}$, \\ Yongliang Shi$^{1,\dag}$, Hao Zhao$^{1}$, Guyue Zhou$^{1}$
\thanks{* Equal contribution. $^{1}$ Institute for AI Industry Research, Tsinghua University; $^{2}$ Xi'an University of Architecture \& Technology; $^{3}$ Beihang University.}%
\thanks{$\dag$ Corresponding author. shiyongliang@air.tsinghua.edu.cn}%
\thanks{Sponsored by Tsinghua-Toyota Joint Research Fund (20223930097).}
}

\usepackage[style=ieee,backend=bibtex,maxbibnames=99]{biblatex}
\AtEveryBibitem{
 \clearlist{address}
 \clearfield{doi}
 \clearfield{note}
 \clearfield{urldate}
 \clearfield{eprint}
 \clearfield{isbn}
 \clearfield{issn}
 \clearlist{location}
 \clearfield{month}
 \clearfield{series}
 
 \ifentrytype{book}{}{
  \clearlist{publisher}
  \clearname{editor}
 }
}

\begin{document}

\maketitle
\thispagestyle{empty}
\pagestyle{empty}


\begin{abstract}
Camera relocalization is a crucial problem in computer vision and robotics. Recent advancements in neural radiance fields (NeRFs) have shown promise in synthesizing photo-realistic images. Several works have utilized NeRFs for refining camera poses, but they do not account for lighting changes that can affect scene appearance and shadow regions, causing a degraded pose optimization process. In this paper, we propose a two-staged pipeline that normalizes images with varying lighting and shadow conditions to improve camera relocalization. We implement our scene representation upon a hash-encoded NeRF which significantly boosts up the pose optimization process. To account for the noisy image gradient computing problem in grid-based NeRFs, we further propose a re-devised truncated dynamic low-pass filter (TDLF) and a numerical gradient averaging technique to smoothen the process. Experimental results on several datasets with varying lighting conditions demonstrate that our method achieves state-of-the-art results in camera relocalization under varying lighting conditions. Code and data will be made publicly available.
\end{abstract}

\section{INTRODUCTION}
Camera relocalization is one of the most important problems in computer vision and robotics. Once we construct an accurate scene map out of a dense collection of images captured by drones or vehicles, we aim to recover the camera pose of given images that are taken inside the reconstructed region, facilitating downstream applications~\cite{slapak_neuralradiance_2023,adamkiewicz_visiononlyrobot_2022}.

Previous methods utilize discriminative networks~\cite{kendall_posenetconvolutional_2015,chen_dfnetenhance_2022,kendall_modellinguncertainty_2016} that perform implicit image feature matching and regress the absolute poses on the given images. These methods can recognize the rough places of the test images but fail to achieve pixel-level accuracy.

Recently, neural radiance fields (NeRFs)~\cite{mildenhall_nerfrepresenting_2020, muller_instantneural_2022,tancik_nerfstudiomodular_2023} have shown their ability to synthesize photo-realistic images. Several works~\cite{yen-chen_inerfinverting_2021,lin_barfbundleadjusting_2021,bian_nopenerfoptimising_2023,jeong_selfcalibratingneural_2021,wang_nerfneural_2022,zhu_latituderobotic_2022,maggio_locnerfmonte_2022} have used NeRF to refine the inaccurate camera pose inputs from a given initialization (may be produced by absolute pose regression (APR) methods~\cite{zhu_latituderobotic_2022}). This line of work optimizes the camera pose by minimizing the photometric error between the rendered image and the observed image. This strategy works well when the lighting conditions remain constant across training and testing sets. Finding the optimal camera pose is then equivalent to rendering the image that can best fit the observation.

\begin{figure}[h]
\centering
\includegraphics[width=0.4\textwidth]{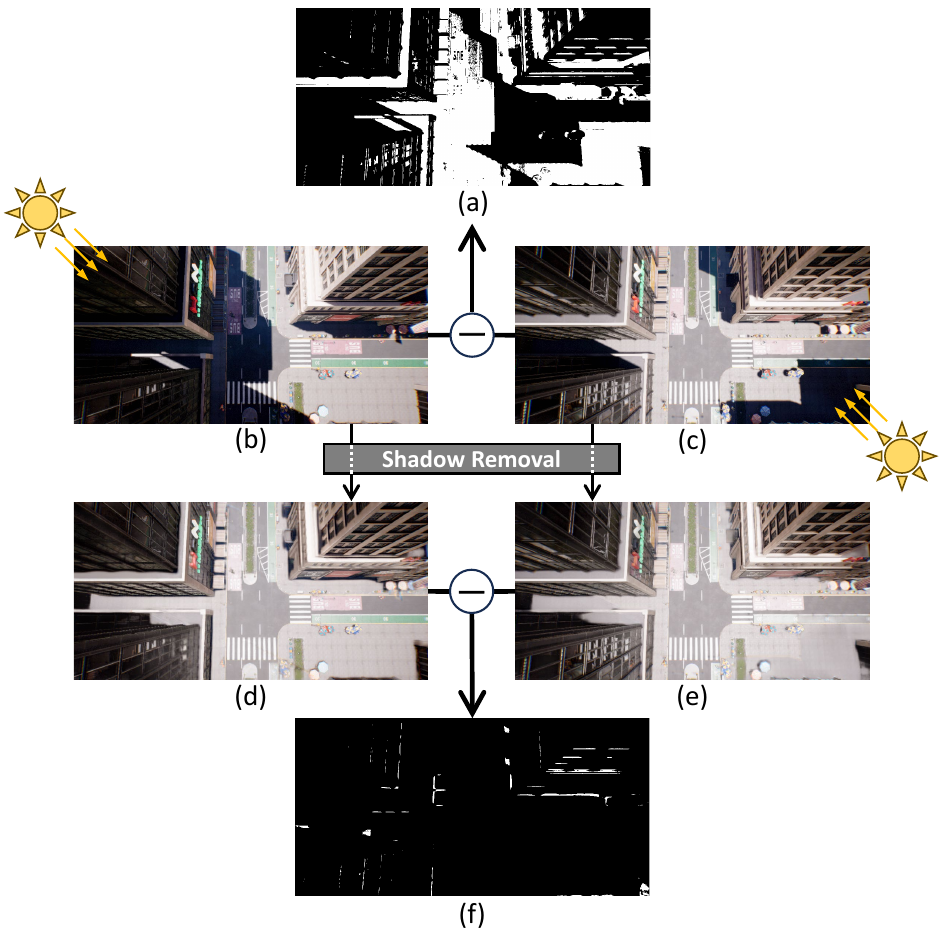}%
\vspace{-2mm}
\caption{Illustration on the negative effects caused by shadows in images. (a) demonstrates the image error in the raw images (b and c) taken under different lighting conditions. Directly optimizing pose via rendering error may cause degraded solutions. Our proposed solution first normalizes the images with a shadow removal network. The shadow-free images and the image error are shown respectively in (d)-(f).}
\label{fig:shadowdifference}
\vspace{-4mm}
\end{figure}

However, this equivalence no longer holds in real-world application settings, where there exist lighting changes that cause the scene appearance and shadow regions to change (as shown in \Figref{fig:shadowdifference}-b,c). Under such circumstances, minimizing the rendering error does not guarantee optimal camera poses. A \textbf{natural and general} solution may be aligning both sets of images by ``normalizing'' the scene lighting. In this paper, we find the key factors that break the equivalence that are needed for pose refining NeRFs and propose a ``normalizer'' that addresses this issue.

As shown in \Figref{fig:shadowdifference}-(a), which illustrates the difference between two images taken at the same camera perspective but with varying lighting conditions, most photometric error occurs in the shadow regions at a given identical camera viewpoint with different lighting conditions. The shadow provides rich image features (e.g., edges, darkness, etc.) when the lighting condition is constant, which helps the convergence but can lead to degraded solutions when lighting changes. In this work, we propose a two-staged pipeline that first aligns both train and test images by a shadow removal module and refines a noisy pose to its global optima robustly in the second stage.

Besides, we implement our neural scene representation with a multi-resolution hash encoding~\cite{muller_instantneural_2022}. This representation scales up the expressiveness and reduces the training time of neural radiance fields. However, combining hash-encoded radiance fields with pose optimization is not trivial. Previous works in NeRF-based pose optimization methods~\cite{lin_barfbundleadjusting_2021,zhu_latituderobotic_2022} rely on the (truncated) dynamic low-pass filter (TDLF) that helps optimize the pose in a coarse-to-fine manner by surpassing the high-frequency domains in the NeRF's positional encoding (PE), which is not applicable in hash-encoded NeRFs.

We notice the similarity between the essence of TDLF used on PE and the multiple resolution levels of hash-coded NeRFs~\cite{muller_instantneural_2022}, where coarse level hash grids correspond to the low-frequency scene structures in positional encoding, while finer level hash grids correspond to the high-frequency scene details in positional encoding. Therefore, we propose to apply a low-pass filter on the weighting to different grid resolution levels. Furthermore, we utilize a numerical gradient averaging technique over the standard autograd operators to encourage smoothened gradient computing.

To sum up, our contributions are:
\begin{enumerate}
    \item We propose a two-staged pipeline that normalizes the images of various lighting and shadow conditions for camera relocalization.
    \item We implement a hash-encoded NeRF for fast training and robust camera pose refinement. A re-devised truncated dynamic low-pass filter and a numerical gradient averaging technique are used to cooperate with the neural scene representation.
    \item We propose a new dataset with varying lighting conditions in training and testing sets. We show that our method achieves state-of-the-art results in camera relocalization. Codes and data will be made publicly available.
\end{enumerate}

\section{Related Work}

\subsection{Camera Relocalization}
Classic visual localization methods can be divided into structure-based and image-based. The former uses 2D keypoints in the image to match with 3D points constructed by Structure-from-Motion (SfM)~\cite{schonberger_structurefrommotionrevisited_2016} to obtain 2D-3D data associations \cite{svarm2016city, taira2018inloc, toft2018semantic, liu2017efficient}. This method gives accurate results but requires the storage of memory-consuming maps and a costly computation. The latter can be realized by image retrieval \cite{gordo2016deep,arandjelovic2016netvlad}, and is often used for place recognition and loop-closure detection, usually only obtaining a rough position. As deep backbone networks demonstrate powerful feature extraction capabilities, SuperPoint \cite{detone2018superpoint} and SiLK \cite{gleize2023silk} learn from self-supervision and outperform classical keypoint detection methods, exhibiting better robustness and generalization ability. 
These days, CNN-based APR methods~\cite{kendall_posenetconvolutional_2015, walch2017image, brahmbhatt2018geometry, xue2020learning} are attracting attention due to their faster speed and better robustness, although the accuracy is not yet sufficiently favorable. PoseNet~\cite{kendall_posenetconvolutional_2015} is the framework of this area, which uses an MLP to regress the camera pose. The later ones improve mainly on the network framework~\cite{loc-deeper, walch2017image, loc-hourglass, loc-apr-transformer} or training strategy~\cite{direct-posenet, loc-glf}.
NeRFs~\cite{mildenhall_nerfrepresenting_2020} can provide photo-realistic images, and they can also be used for APR tasks. LENS~\cite{moreau2022lens} uses a NeRF-W network~\cite{martin-brualla_nerfwild_2021} to synthesize realistic and geometry-consistent images as data augmentation during training and achieves higher localization accuracy. Subsequently, DFNet~\cite{chen_dfnetenhance_2022} introduces an online synthetic data generation scheme and proposes a network that extracts domain invariant features in order to reduce the domain gap between synthetic and real images. LATITUDE~\cite{zhu_latituderobotic_2022} combines an APR with a pose optimizer to localize in the city but does not take into account the shadow area caused by the scene lighting changes which are very common in outdoor scenes. We address this problem with a shadow-removal pipeline that normalizes all the input images from various lighting conditions.

\subsection{Optimization-based Relocalization}
Differentiable rendering methods, such as NeRF, make it possible to recover the camera pose by back-propagating the scene representation. iNeRF~\cite{yen-chen_inerfinverting_2021} proposes the first framework that uses a reconstructed NeRF model to estimate the camera pose. To eliminate the negative impact of positional encoding on pose registration, BARF~\cite{lin_barfbundleadjusting_2021} presents a coarse-to-fine strategy to jointly optimize scene representations and camera poses. Subsequently, \cite{chng2022gaussian} uses Gaussian-MLPs to simplify the process of solving the joint task of scene representation and pose optimization. However, the above methods are restricted to small indoor scenes. LATITUDE~\cite{zhu_latituderobotic_2022} introduces a two-stage localization mechanism to solve the global localization problem of large-scale scenes. To avoid local optimum, it applies a truncated dynamic low-pass filter during the optimization stage. However, a common problem with the above methods is that they are all MLP-based and still take an extensive amount of time. Many recent works have focused on accelerating the NeRF training, including but not limited to~\cite{fridovich-keil_plenoxelsradiance_2022, sun_directvoxel_2022, muller_instantneural_2022}. InstantNGP~\cite{muller_instantneural_2022} uses grid sampling and multi-resolution hash encoding to speed up the convergence greatly. In this work, we extend the current pose optimization scheme to be combined with hash-encoded NeRFs.

\newcommand{\testimage}{I^{(l')}}
\newcommand{\nerfnetwork}{\mathcal{F}}
\newcommand{\shadownet}{\mathcal{N}_\text{shadow}}

\section{Formulation \& Preliminaries}
Our goal is to retrieve the accurate camera pose $T(\testimage)$ for a given image $\testimage$ with its corresponding lighting condition $l'$ in a pre-reconstructed neural radiance field $\nerfnetwork^{(l)}$ with lighting condition $l$. This problem is well addressed when $l=l'$ in existing literature such as LATITUDE~\cite{zhu_latituderobotic_2022}: first, we pass the test image $\testimage$ through an absolute pose regressor (APR) network, obtaining an initial guess for the pose, denoted as $T_0$. This initial pose may be noisy due to the implicit nature of CNN-based APR methods. Next, we optimize the inaccurate pose prediction iteratively in a stochastic gradient descent (SGD) manner via:

\begin{equation}
\label{eq:sgd}
    T_{i+1} = T_{i} - \alpha\cdot \frac{\partial \mathcal{L}(I - \nerfnetwork(T_{i}))}{\partial T_{i}},
\end{equation}
where $\alpha$ is the learning rate and $\mathcal{L}(I - \nerfnetwork(T_{i}))$ is the rendering error between the test image and the rendered image by the NeRF network $\nerfnetwork$ at pose $T_i$.

This optimization pipeline leads to the optimal solution $T^*$ when the lighting condition $l$ remains constant at most times, as described as an equivalence shown in \eqref{eq:equiv}. However, when $l\neq l'$, the optimization scheme in \eqref{eq:sgd} cannot lead to an optimal solution since $\mathcal{L}(\testimage - \nerfnetwork^{(l)}(T^*))$ may not be the minimal error anymore.

\begin{equation}
    \label{eq:equiv}
    \mathcal{L}(I - \nerfnetwork(T^*))=0.
\end{equation}

This photometric inconsistency causes the sensitive feature matching steps of \eqref{eq:sgd} to fail. This key observation then leads us to a natural solution of aligning all images to a ``normalized'' lighting condition $l_0$ by a normalizer $\mathcal{N}(I^{(l_*)}) = I^{(l_0)}$. DFNet~\cite{chen_dfnetenhance_2022} proposes to align the images with a histogram-assisted NeRF. However, as shown in \Figref{fig:shadowdifference}, the photometric error caused by lighting changes is primarily due to shadow changes.

Our two-staged pipeline addresses this issue by aligning the training set images $\{I_k^{(l)}\}$ and the test image $\testimage$ with a shadow removal network $\shadownet$. In the first stage, we train a neural radiance field $\nerfnetwork^{(l_0)}$ with normalized images. Then, in the second stage, we optimize the test camera pose by minimizing $\mathcal{L}(\shadownet(\testimage) - \nerfnetwork^{(l_0)}(T_i))$.

\begin{figure*}[t]
\centering
\includegraphics[width=0.98\textwidth]{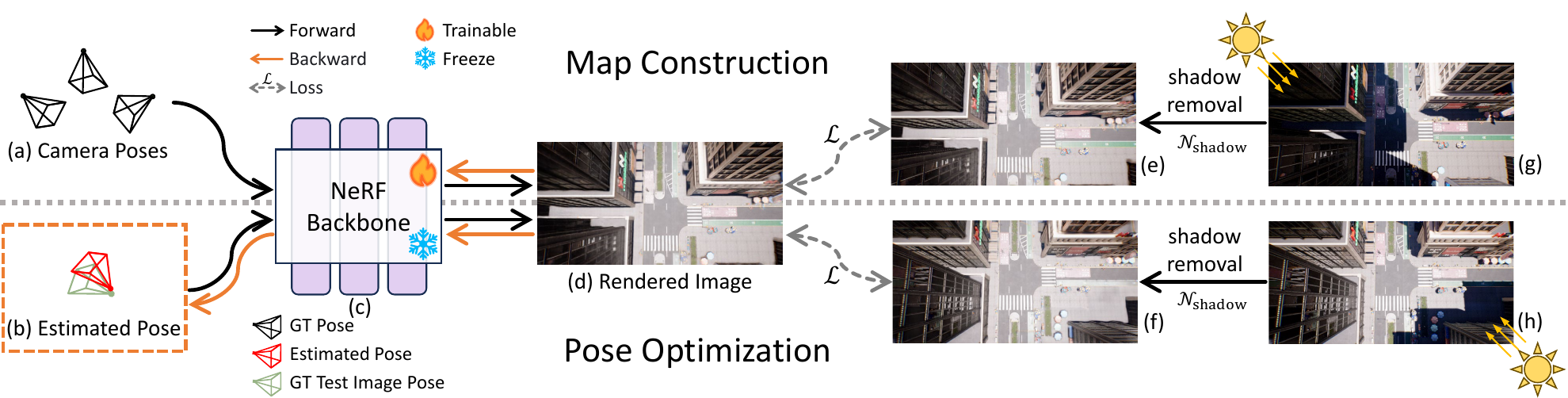}%
\caption{\textbf{Pipeline}. \textbf{Top}: In the map construction process, we fit a hash-encoded NeRF onto a set of shadow-free images. \textbf{Bottom}: Once the NeRF is trained, we can recover the camera pose for any given test image $\testimage$ by first processing the image with the same shadow removal network $\shadownet$ as used in the training stage and refine the initial pose recursively with the NeRF network fixed.}
\label{fig:pipeline}
\end{figure*}
\vspace{-3mm}
\newcommand{\shadowdetachnet}{\mathcal{D}_\text{shadow}}
\newcommand{\shadowremovalnet}{\mathcal{R}_\text{shadow}}

\section{Method}

\subsection{Pipeline}

As illustrated in \Figref{fig:pipeline}, our proposed method is structured as a two-stage pipeline: 1. We reconstruct the scene map $\nerfnetwork$ (top row in \Figref{fig:pipeline}) with normalized lighting conditions $l_0$ by pre-processing the training images with a shadow removal network $\shadownet$; 2. We find the accurate camera pose $T$ of the test image $\testimage$ (bottom row in \Figref{fig:pipeline}). For the scene map, we construct a three-dimensional neural scene map based on a multi-resolution hash grid using a set of posed RGB images.

It's noteworthy that the images used during the map construction and pose optimization stages contain various shadows. Since the shadow variations as one major manifestation of photometric inconsistency (as shown in \Figref{fig:shadowdifference}-a), we employ the shadow removal network, denoted as $\shadownet$, for images used in NeRF map construction and pose optimization, to obtain ``normalized'' shadow-free image components, further satisfying the equivalence in \eqref{eq:equiv}.

\textbf{Shadow Removal:} In our image shadow removal process, there are two distinct steps: the first is shadow detection, and the second is shadow removal reliant on the shadow mask produced in the first step. For shadow detection procedure $\shadowdetachnet$, we utilized MTMT~\cite{chen20MTMT}, which segments shadow in the input images $I^{(l_{*})}$, yielding a shadow region mask $M$. For the shadow removal process $\shadowremovalnet$, we employed a Transformer-based network following \cite{guo2023shadowformer}. By inputting the previously obtained mask and the input image, it associates the shadowed regions with non-shadowed regions to remove shadows within the masked area, producing a shadow-free image. The shadow removal process can be expressed as:
\begin{equation}
    I^{(l_0)}=\shadownet(I^{(l_{*})}) = \shadowremovalnet (I^{(l_{*})}, \shadowdetachnet (I^{(l_{*})})).
\end{equation}

\textbf{Map Construction:} 
In this section, we detailed our hash-encoded scene representation $\nerfnetwork$. The scene representation network is approximated with a multi-resolution hash grid, along with a shallow MLP decoder. Given a 3D position $\mathbf{x}\in\mathbb{R}^3$. We query the hash grid in each of the $L$ resolution levels to obtain the hash feature $\{he_k(\textbf{x})\}_{k=1}^L$. 

The hash encoding $he_k(\mathbf{x})$ at the $k^{th}$ resolution level is derived from tri-linear interpolation on the 8 neighboring grid points around the queried position:
\begin{equation}
\label{eq:hashvalue}
    he_k(\mathbf{x})=\mathtt{interp}_k\left(\left(\bigoplus_{i=1}^{3} x_{i} \pi_{i}\right) \bmod T\right),
\end{equation}
where $\mathtt{interp}_k(\cdot)$ denotes the tri-linear interpolation operator in the $k^{th}$ resolution level grid, $\pi_i$ and $T$ are the parameters of the hash function.

For each resolution grid, we obtain an $F$-dimensional feature vector. Subsequently, the obtained $L$ feature vectors are sequentially concatenated:
\begin{equation}
\label{eq:hashencodingL}
    HE(\mathbf{x})=(he_1(\mathbf{x}), he_2(\mathbf{x}), \cdots , he_L(\mathbf{x})).
\end{equation}

We employ a shallow MLP to predict the sampled point color and density from the extracted hash feature $HE(\textbf{x})$ and direction $\mathbf{d}\in\mathbb{R}^3$ to the volume density $\sigma$ of the sample point and its color $\mathbf{c}$. We follow the approach of Instant-NGP~\cite{muller_instantneural_2022} and utilize spherical harmonics $SH(\cdot)$ for direction encoding. The forwarding pass of our NeRF network can be described as:
\begin{equation}
\label{eq:nerfmlp}
    \mathbf{c}, \mathbf{\sigma} = \nerfnetwork(\mathbf{x}, \mathbf{d}) = \mathrm{MLP}(HE(\mathbf{x}), SH(\mathbf{d})).
\end{equation}

Subsequently, the rendered image $\hat{I}^{(l_0)}$ is obtained using volume rendering. To achieve the alignment of training images from their original lighting condition to the shadow-free lighting condition $l_0$, the loss function we employ during the NeRF training phase is:
\begin{equation}
\label{eq:nerftrainloss}
    \mathcal{L}=\sum_{\mathbf{r} \in \mathcal{R}} \left| \hat{I}^{(l_0)}(\mathbf{r})-\shadownet(I^{(l_{*})}(\mathbf{r})) \right|_1,
\end{equation}

\noindent
where $\mathcal{R}$ represents a batch of rays obtained through sampling during the training process, $\mathbf{r}$ is a ray from the set $\mathcal{R}$, and $I^{(l_{*})}$ denotes the images from the training set with an arbitrary lighting condition $l_*$.

This loss function compels the network's rendering output to closely resemble the shadow-free training images. Consequently, the constructed map represents a normalized scene devoid of shadows, laying a solid foundation for pose optimization. During the pose optimization stage in section \ref{subsec:poseoptimization}, the same shadow removal model is utilized to remove shadows from the test images, ensuring that both the map and the images used for pose optimization are aligned to $l_0$.

\begin{figure}[!h]
\centering
\includegraphics[width=0.48\textwidth]{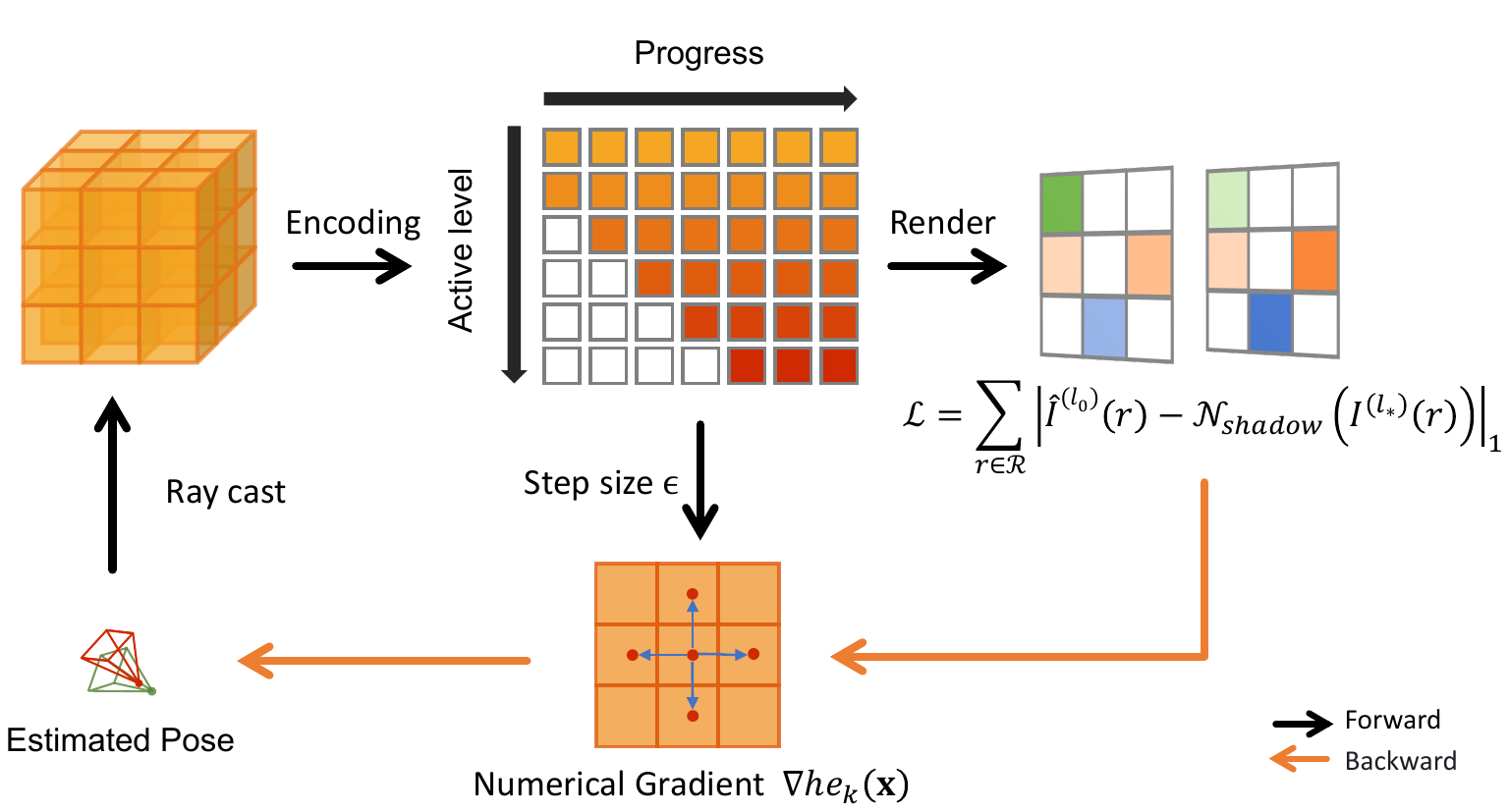}%
\caption{Illustration on the \textbf{re-devised TDLF} and the \textbf{numerical gradient averaging} techniques. During the pose optimization stage, we filter the extracted feature on its frequency domain by using a set of closed-form weighting parameters. When updating the estimated pose with SGD, we use numerical gradient averaging over the standard Autograd operators to ensure a smooth optimization process.}
\label{fig:poseoptimize}
\end{figure}

\subsection{Pose Optimization}
\label{subsec:poseoptimization}

Pose optimization in NeRFs involves refining the camera's pose to align the virtual and real-world scenes accurately. Given a shadow-free image $I^{(l_0)}$ and corresponding initial pose guess $\hat{T}_0$, the goal is to minimize the photometric loss between the observed image and the rendered image $\hat{I}$ which is obtained from camera pose. We optimize the camera pose on the manifold and express the parameter of the camera extrinsic as $\xi \in \mathfrak{se}(3)$, the optimized pose is obtained by equation \ref{eq:poseobtain} as:
\begin{equation}
\label{eq:poseobtain}
    \hat{T}=\exp(\hat{\xi})\hat{T}_0.
\end{equation}

To optimize the pose parameters, we use the same loss function as in the map reconstruction stage (equation~\ref{eq:nerftrainloss}).

However, trivially minimizing the loss function may lead to sub-optimal results. We further propose a coarse-to-fine optimization strategy that is in the same spirit as the dynamic low-pass filters proposed in \cite{lin_barfbundleadjusting_2021,zhu_latituderobotic_2022} and a numerical gradient averaging technique which smoothens the optimization process, as shown in 
Fig.~\ref{fig:poseoptimize}.

\textbf{Coarse-to-fine Optimization: } Since the input images are rich in high-frequency signals, using gradient descent to minimize photometric loss can lead to local minima rather than finding the global minimum. A coarse-to-fine smoothing technique (referred to as the truncated dynamic low-pass filter, or the TDLF) had previously been proposed~\cite{lin_barfbundleadjusting_2021,zhu_latituderobotic_2022}, in which the feature of position encoding is used to separate the high- and low-frequency image components of the scene and the high-frequency details are smoothed out in the early stage of pose estimation. The TDLF are implemented by suppressing the high-frequency components of the positional encoding (PE). Since PE is no longer applicable in our grid-based scene representation, it is not trivial to utilize TDLF in our setting. In this paper, we extend the TDLF to grid-based NeRFs by appending weighting parameters on the extracted feature from each resolution level as:
\begin{equation}
\label{eq:hashencodingFilter}
    HE(\mathbf{x})=(\omega_1(\alpha) he_1(\mathbf{x}), \cdots, \omega_L(\alpha) he_L(\mathbf{x})),
\end{equation}
where $\alpha \in [0, L]$ is a function of the optimization progress. While a pre-trained NeRFs model is given, rendering with too few resolution levels activated can lead to invalid outputs, we set an appropriate initial $\alpha$ to avoid this problem by setting $\alpha = \min(\alpha_0 + \text{progress}, 1)L$. The weighting parameter for level $k$ is set to:
\begin{equation}
\label{eq:hashencodingWeight}
w_k(\alpha)=\begin{cases}0&\ \mathrm{if}\ \alpha<k\\\frac{1-\cos((\alpha-k)\pi)}2&\ \mathrm{if}\ 0\leq\alpha-k<1\\1&\ \mathrm{if}\ \alpha-k\geq1\end{cases}.
\end{equation}

\textbf{Numerical gradient averaging: } The numerical gradient averaging technique aims to solve the derivative discontinuity problem caused by hash encoding. From the loss function described in equation \ref{eq:nerftrainloss} described previously, we can derive the steepest descent direction for pose variables:
\begin{equation}
\label{eq:posejacobian}
    \mathbf{J}(\mathbf{u}; \xi)=\sum_{i=1}^N\frac{\partial g(\mathbf{y}_1,\ldots,\mathbf{y}_N)}{\partial\mathbf{y}_i}\frac{\partial\mathbf{y}_i}{\partial\mathbf{x}_i(T)}\frac{\partial\mathcal{W}(T)}{\partial \xi},
\end{equation}
where $g$ represents the volume rendering process, $\mathbf{x}_i$ is the 3D point along the sampling ray with corresponding hash encoding $\mathbf{y}_i$ and $\mathcal{W}$ represents the ray casting process.

Hash encoding has localized derivatives, meaning that when points cross grid cell boundaries, the resulting hash entries will change discontinuously, resulting in disrupt changes in $\frac{\partial\mathbf{y}_i}{\partial\mathbf{x}_i(T)}$. This further blocks the feature-matching process that is vital for the pose optimization process. Inspired by recent work in neural 3D reconstruction~\cite{li_neuralangelohighfidelity_2023}, we introduce a numerical gradient averaging technique to allow for feature sharing among neighboring grid cells.

To compute the numerical gradient, we sample a set of additional points around the queried points. To be specific, given a 3D point $\mathbf{x}_{i}=(x_{i},y_{i},z_{i})$ on a sampling ray, we need to sample two points along each axis of canonical coordinates around $\mathbf{x}_{i}$ with a step size of $\epsilon$. For example, the $x$-component of $\nabla he_k(\mathbf{x}_i)$ can be found as
\begin{equation}
\label{eq:numericalgrad}
    \nabla_x he_k(\mathbf{x}_i)=\frac{he_k\left(\mathbf{x}_i+\boldsymbol{\epsilon}_x\right)-he_k\left(\mathbf{x}_i-\boldsymbol{\epsilon}_x\right)}{2\epsilon},
\end{equation}
where $\epsilon_x=[\epsilon,0,0]$, $\epsilon$ is the inverse of the resolution of the currently activated level $L_\text{act}$ which is controlled by $\alpha$ with $L_\text{act} = \lceil \alpha \rceil$. In total, 6 additional points are sampled to compute the full numerical gradient.

Although the numerical gradient would be equivalent to the analytical gradient if the step size is smaller than the grid size, the numerical gradient can still smooth gradients when $\mathbf{x}_i$ near the borders as $\mathbf{x}_i \pm \boldsymbol{\epsilon}$ can move across the grids.
\section{Experiments and Analysis}
\subsection{Dataset}
We evaluate our method using the New York scene in our proposed Shadow Urban Minimum Altitude Dataset (SUMAD) and partial scenes from the public dataset NeRF-OSR~\cite{rudnev_nerfoutdoor_2022}. The SUMAD is a virtual-scene dataset made by the simulator AirSim~\cite{airsim2017fsr} built on top of the Unreal Engine. To replicate various real-world shadowing scenarios, we manipulate the direction and position of the light source within Unreal Engine, creating three distinct types of lighting conditions for each scene. Further, to enhance data realism, we incorporate two authentic city scene models within Unreal Engine, faithfully replicating urban environments resembling New York and San Francisco. It is worth noting that we are the first to release a dataset featuring diverse shadowing scenarios within large-scale city environments. The NeRF-OSR data is a set of real-world data captures that contain different lighting conditions.

\subsection{Implementation Details}
We used a hash grid configuration following InstantNGP~\cite{muller_instantneural_2022}. We train the scene for 100,000 iterations. The base MLP consists of 1 hidden layer of 64 units and output 16 channels which together with the direction encoded by the harmonic function forms the input to the head MLP, which consists of 2 hidden layers of 64 units. We resize the images to 960 × 480 pixels and randomly cast rays during the training steps, with constrain on the amount of 3D sampling points to be limited to $1<<18$. We use an Adam optimizer with an initial learning rate of $1 \times 10^{-2}$ decaying exponentially to $1 \times 10^{-4}$ for scene reconstruction and $1.2 \times 10^{-2}$ to $1.2 \times 10^{-3}$ for pose optimize, while a scaler is applied to magnify loss by $2^{10}$. 

In order to implement the coarse-to-fine strategy in the hash grid encoding and numerical gradient, we add a deterministic layer after hash grid encoding, which sets the active level and step size of the numerical gradient. The active level is initially set to 8 in the experiments. All experiments are conducted on a Linux system with an NVIDIA RTX3090 GPU with 24GB of memory, and 1000 iterations are run for each optimization.

\begin{table*}[t]
\vspace{5mm}
\renewcommand\arraystretch{1.2}
\centering
\caption{Quantitative results with baseline methods.}
\vspace{-2mm}
\label{tab:maintable}
\resizebox{0.78\textwidth}{!}{
\begin{tabular}{l|cc|ccc|c}
\hline
\begin{tabular}[c]{@{}c@{}}Methods\end{tabular} &
  \begin{tabular}[c]{@{}c@{}}Translation Error(m)\end{tabular} &
  \begin{tabular}[c]{@{}c@{}}Rotation Error($^\circ$)\end{tabular} & 
  \begin{tabular}[c]{@{}c@{}}PSNR$\uparrow$ \end{tabular} & 
  \begin{tabular}[c]{@{}c@{}}SSIM$\uparrow$ \end{tabular}&
  \begin{tabular}[c]{@{}c@{}}LPIPS$\downarrow$ \end{tabular}& 
  \begin{tabular}[c]{@{}c@{}}Time\\ Train / Inference \end{tabular}\\ 
  \hline
 DFNet~\cite{chen_dfnetenhance_2022}& 491.457 & 30.315 &- & -&- & 40h / $<1$s \\ 
 DFNet ($+\shadownet$) & 6.002 & 28.654 &- &- &- & 40h / $<1$s \\ 
  \hline
 LATITUDE~\cite{zhu_latituderobotic_2022}& 12.174 & 11.803 & 9.260 & 0.207 & 0.587 & 10h / 5.5m \\ 
 LATITUDE ($+\shadownet$) & 0.094 & 0.534 & 24.028 & 0.758 & \textbf{0.195} & 10h / 5.5m\\  
iNeRF+iNGP~\cite{yen-chen_inerfinverting_2021,muller_instantneural_2022} & 8.697 & 8.832 & 9.766 & 0.248 & 0.635 & 1.5h / 27s \\
 \textbf{Ours} ($-\shadownet$)  & 6.747 & 8.398 & 9.831 & 0.219 & 0.636 & 1.5h / 55s \\
 \textbf{Ours} & \textbf{0.091} & \textbf{0.106} & \textbf{25.799} & \textbf{0.826} & 0.253 & 1.5h / 55s \\
 \hline
\end{tabular}
}
\end{table*}
\subsection{Results}

The quantitative evaluations on the SUMAD dataset are shown in Table \ref{tab:maintable}. During experiments, we introduce a 4-meter translation error parallel to the trajectory forward direction to the ground truth pose as an initial pose for all methods and evaluate all of the methods on image synthesis results and the translation/rotation errors on the recovered poses. Note that the series of experiments with DFNet are not applicable for introducing initial pose errors and for evaluating image synthesis qualities. The image synthesis metrics (PSNR, SSIM, and LPIPS~\cite{zhang_unreasonableeffectiveness_2018}) are evaluated on the shadow-free images no matter whether the shadow removal process is used or not for fair comparisons.

The baseline methods evaluated here are: 1. DFNet~\cite{chen_dfnetenhance_2022} trained on the raw images and evaluated on test images with different lighting conditions, which causes a degraded solution in regressing the absolute position of the test images; 2. DFNet variant that is trained and evaluated on the shadow-free images shows a significant improvement in localization accuracy (from 491m to 6m); 3/4. LATITUDE~\cite{zhu_latituderobotic_2022} trained on raw/shadow-free images; 5. Performing iNeRF-like~\cite{yen-chen_inerfinverting_2021} direct pose refinement on an InstantNGP~\cite{muller_instantneural_2022}; 6/7. Our proposed method with and without shadow removal process.

As can be inferred from the comparisons between shadow-free and unprocessed counterparts, the shadow removal process significantly improves the image synthesis quality and the relocalization accuracy, no matter the scene representation used (DFNet~\cite{chen_dfnetenhance_2022}, LATITUDE~\cite{zhu_latituderobotic_2022}, and our proposed hash-encode NeRF). Among all the evaluated baseline methods, our proposed two-staged pipeline achieves the best image synthesis quality, the most accurate poses recovered, and the fastest overall time used for training and evaluation.

\subsection{Ablation Study}

\begin{table}[!t]
\renewcommand\arraystretch{1.2}
\centering
\caption{Ablation study with different initial translation errors.}
\label{tab:transerror}
\resizebox{0.8\linewidth}{!}{
\begin{tabular}{cccccc}
\hline
\begin{tabular}[c]{@{}c@{}}Initial\\ Error(m)\end{tabular} &
  \begin{tabular}[c]{@{}c@{}}Numerical Gradient\\ Averaging\end{tabular} &
  \begin{tabular}[c]{@{}c@{}}TDLF\end{tabular} &
  \begin{tabular}[c]{@{}c@{}}Translation\\ Error(m)\end{tabular} &
  \begin{tabular}[c]{@{}c@{}}Rotation\\ Error(${}^{\circ}$)\end{tabular} \\
  \hline
\multirow{4}{*}{4}  & $\bm{\times}$  & $\bm{\times}$  & 0.12          & 0.13          \\
                    & $\bm{\times}$ & \checkmark  & 0.12          & 0.13          \\
                    & \checkmark & $\bm{\times}$  & 0.09          & 0.11          \\
                    & \checkmark & \checkmark & \textbf{0.09} & \textbf{0.11} \\ \hline
\multirow{4}{*}{8}  & $\bm{\times}$  & $\bm{\times}$  & 3.42          & 1.22          \\
                    & $\bm{\times}$ & \checkmark  & 1.18          & 0.27          \\
                    & \checkmark & $\bm{\times}$  & 0.09          & 0.10          \\
                    & \checkmark & \checkmark & \textbf{0.09} & \textbf{0.10} \\ \hline
\multirow{4}{*}{12} & $\bm{\times}$  & $\bm{\times}$  & 4.20          & 1.44          \\
                    & $\bm{\times}$ & \checkmark  & 4.20          & 1.20          \\
                    & \checkmark & $\bm{\times}$  & 1.85          & 1.69          \\
                    & \checkmark & \checkmark & \textbf{0.10} & \textbf{0.11} \\ \hline
\multirow{4}{*}{16} & $\bm{\times}$  & $\bm{\times}$  & 11.05          & 4.68          \\
                    & $\bm{\times}$ & \checkmark  & 10.09          & 4.37          \\
                    & \checkmark & $\bm{\times}$  & 3.71          & 4.13          \\
                    & \checkmark & \checkmark & \textbf{0.07} & \textbf{0.09} \\ \hline
\end{tabular}
}
\end{table}

\begin{figure}[!t]
\vspace{1mm}
\centering
\includegraphics[width=0.48\textwidth]{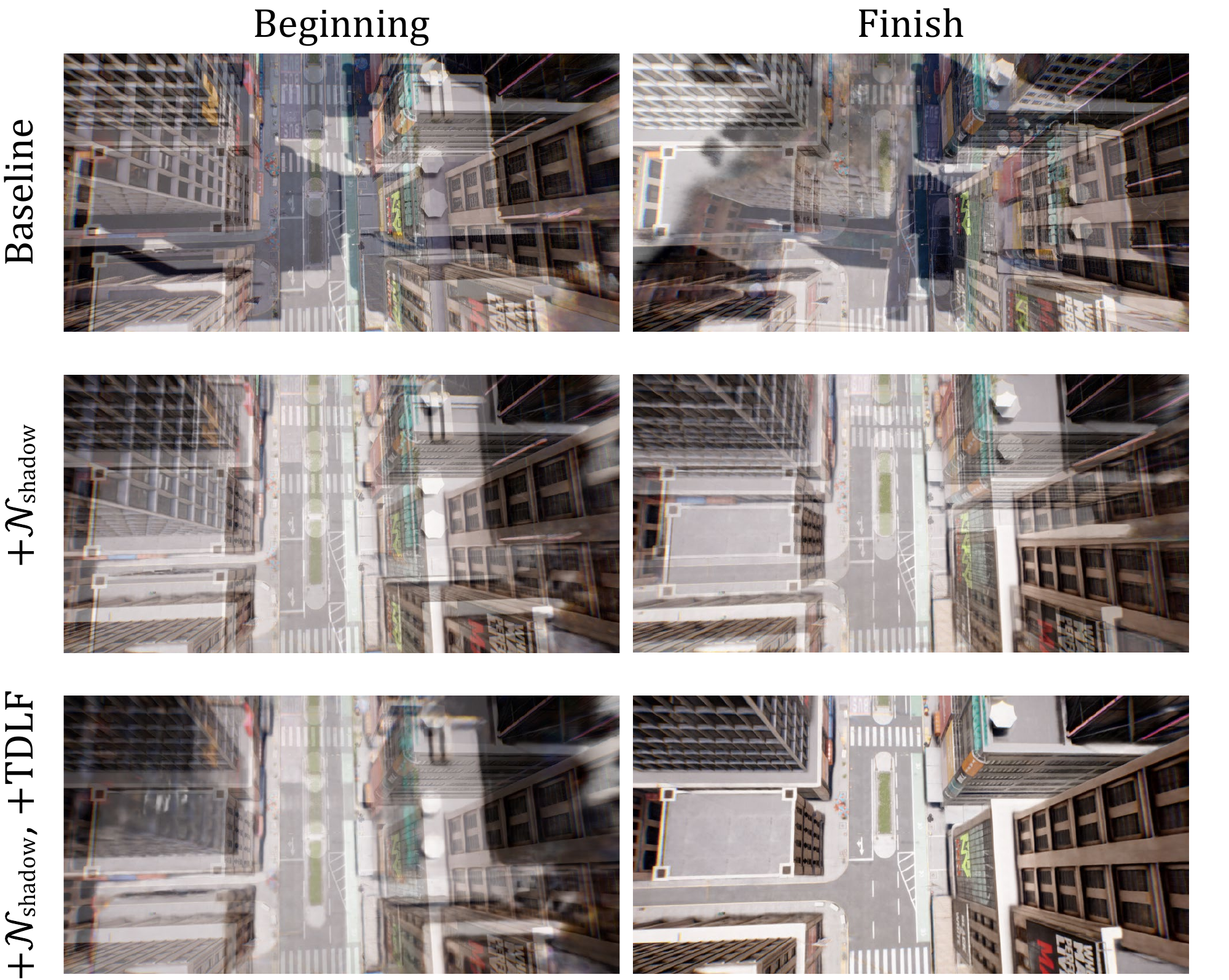}%
\caption{We show the optimization process of our methods with different design choices. The rendered and observed images are stacked to demonstrate both rendering quality and pose accuracy.}
\label{fig:ablationfig}
\end{figure}

\begin{table}[!t]
\renewcommand\arraystretch{1.2}
\centering
\caption{Ablation study with different initial rotation error.}
\label{tab:roterror}
\resizebox{0.8\linewidth}{!}{
\begin{tabular}{ccccc}
\hline
\begin{tabular}[c]{@{}c@{}}Initial\\ Error(${}^{\circ}$)\end{tabular} &
  \begin{tabular}[c]{@{}c@{}}Numerical Gradient\\ Averaging\end{tabular} &
  \begin{tabular}[c]{@{}c@{}}TDLF\end{tabular} &
  \begin{tabular}[c]{@{}c@{}}Translation\\ Error(m)\end{tabular} &
  \begin{tabular}[c]{@{}c@{}}Rotation\\ Error(${}^{\circ}$)\end{tabular} \\
  \hline
\multirow{4}{*}{4}  & $\bm{\times}$  & $\bm{\times}$  & 0.11          & 0.13          \\
                    & $\bm{\times}$ & \checkmark  & 0.10          & 0.11          \\
                    & \checkmark & $\bm{\times}$  & 0.10          & 0.10          \\
                    & \checkmark & \checkmark & \textbf{0.10} & \textbf{0.10} \\ \hline
\multirow{4}{*}{8}  & $\bm{\times}$  & $\bm{\times}$  & 0.32          & 0.22          \\
                    & $\bm{\times}$ & \checkmark  & 0.10          & 0.11          \\
                    & \checkmark & $\bm{\times}$  & 2.58          & 3.37          \\
                    & \checkmark & \checkmark & \textbf{0.10} & \textbf{0.10} \\ \hline
\multirow{4}{*}{12} & $\bm{\times}$  & $\bm{\times}$  & 2.34          & 1.86         \\
                    & $\bm{\times}$ & \checkmark  & 3.18          & 3.55          \\
                    & \checkmark & $\bm{\times}$  & 3.18          & 2.79          \\
                    & \checkmark & \checkmark & \textbf{1.06} & \textbf{0.45} \\ \hline
\multirow{4}{*}{16} & $\bm{\times}$  & $\bm{\times}$  & 7.75          & 5.41          \\
                    & $\bm{\times}$ & \checkmark  & 9.07          & 5.92          \\
                    & \checkmark & $\bm{\times}$  & 5.22          & 3.69          \\
                    & \checkmark & \checkmark & \textbf{4.34} & \textbf{3.60} \\ \hline
\end{tabular}
}
\end{table}

Our method uses a coarse-to-fine strategy and numerical gradient averaging technique to achieve a more accurate and robust pose optimization process. To evaluate the effects of our proposed improvements, we perform an ablation study that quantitatively analyzes how the error tolerance and convergence accuracy of pose optimization change when the improvements are applied or not.

\begin{figure}[!t]
\centering
\includegraphics[width=0.475\textwidth]{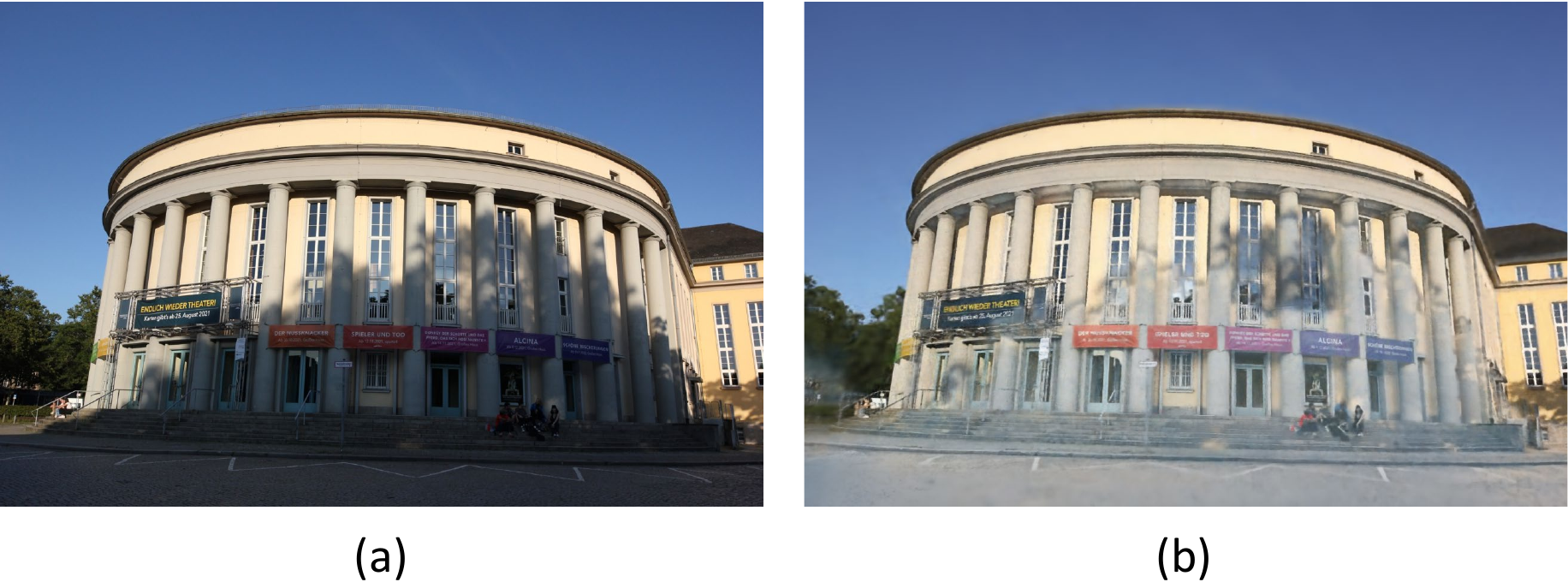}%
\caption{Result of our method optimizing pose on the NeRF-OSR dataset. \ \   (a) Original image from dataset at the ground truth pose. (b) Images rendered with optimized noisy pose.}
\label{fig:nerfosrfig}
\end{figure}

Experiments are carried out at six randomly selected positions in the SUMAD dataset, with initial translation or rotation errors of the same size introduced at each position. The average error results are recorded in table \ref{tab:transerror} and table \ref{tab:roterror}. The results presented in the tables demonstrate that our proposed method exhibits the greatest robustness to both translation and rotation perturbations, and also achieves the highest convergence accuracy out of all methods tested. The results validate the effectiveness of incorporating the coarse-to-fine strategy into state estimation methods based on Grid-based NeRFs. The coarse-to-fine strategy can help the pose optimization to escape from local minima generated by the high-frequency information, as shown in 
Fig.~\ref{fig:ablationfig}. Besides, numerical gradient also shows its superiority compared to analytical gradient in the context of hash grid encoding even without the coarse-to-fine strategy, which is consistent with what is expected in section \ref{subsec:poseoptimization}. We also conducted tests using our method on partial scenes from NeRF-OSR and obtained relatively good results, as shown in 
Fig.~\ref{fig:nerfosrfig}.

\section{CONCLUSIONS}
In this paper, we address the challenge of camera pose refinement under varying lighting conditions. We've pinpointed shadow differences as a principal source of photometric errors, highlighting the necessity of consistent lighting for accurate pose refinement in NeRFs.
Our proposed two-staged pipeline ensures images are normalized irrespective of shadow and lighting variations. By introducing a shadow removal module, we've successfully bridged the photometric discrepancies between images. Further, the integration of multi-resolution hash encoding with our neural scene representation has not only amplified its expressiveness but has also substantially expedited the training process. We've ingeniously combined this with pose optimization, offering a method that is both accurate and efficient.
Our method achieves state-of-the-art results on our proposed dataset and other public dataset.



\printbibliography

\end{document}